\pdfoutput=1

\documentclass[11pt]{article}

\usepackage[]{EMNLP2022}

\usepackage{times}
\usepackage{latexsym}

\usepackage[T1]{fontenc}

\usepackage[utf8]{inputenc}

\usepackage{microtype}

\usepackage{inconsolata}

\usepackage{xspace,mfirstuc,tabulary}

\usepackage{amsmath}

\usepackage{cleveref}
\crefformat{section}{\S#2#1#3}
\crefformat{subsection}{\S#2#1#3}
\crefformat{subsubsection}{\S#2#1#3}
\crefrangeformat{section}{\S\S#3#1#4 to~#5#2#6}
\crefmultiformat{section}{\S\S#2#1#3}{ and~#2#1#3}{, #2#1#3}{ and~#2#1#3}
\usepackage{refstyle}
\Crefformat{figure}{#2Figure~#1#3}
\Crefmultiformat{figure}{Figs.~#2#1#3}{ and~#2#1#3}{, #2#1#3}{ and~#2#1#3}
\Crefformat{table}{#2Table~#1#3}
\Crefmultiformat{table}{Tabs.~#2#1#3}{ and~#2#1#3}{, #2#1#3}{ and~#2#1#3}
\Crefformat{appendix}{Appx.~\S#2#1#3}
\crefformat{algorithm}{Alg.~#2#1#3}

\usepackage[utf8]{inputenc}

\usepackage{microtype}

\usepackage{graphicx}
\usepackage{booktabs}
\usepackage{bm}
\usepackage{amsmath,amssymb}
\usepackage{color, colortbl}
\usepackage{todonotes}
\title{Revisiting Parameter-Efficient Tuning: Are We Really There Yet?}

\author{
  Guanzheng Chen$^{1}$,
  Fangyu Liu$^{2}$,
  Zaiqiao Meng$^{3}$,
  Shangsong Liang$^{1,4,}$\Thanks{~Corresponding author.}
  \\
$^1$Sun Yat-sen University \ \
  $^2$University of Cambridge \ \
$^3$University of Glasgow
  \\
  $^4$Mohamed bin Zayed University of Artificial Intelligence
  \\
  \small \texttt{guanzzh.chen@gmail.com, fl399@cam.ac.uk}
  \\
  \small \texttt{zaiqiao.meng@glasgow.ac.uk, liangshangsong@gmail.com}
}

\begin{document}
\maketitle
\begin{abstract}
Parameter-Efficient Tuning (PETuning) methods have been deemed by many as the new paradigm for using pretrained language models (PLMs). By tuning just a fraction amount of parameters comparing to full model finetuning, PETuning methods claim to have achieved performance on par with or even better than finetuning. In this work, we take a step back and re-examine these PETuning methods by conducting the first comprehensive investigation into the training and evaluation of them. We found the problematic validation and testing practice in current studies, when accompanied by the instability nature of PETuning methods, has led to unreliable conclusions. When being compared under a truly fair evaluation protocol, PETuning cannot yield consistently competitive performance while finetuning remains to be the best-performing method in medium- and high-resource settings. 
We delve deeper into the cause of the instability and observed that the number of trainable parameters and training iterations are two main factors: reducing trainable parameters and prolonging training iterations may lead to higher stability in PETuning methods.\footnote{Our code is available at \url{https://github.com/guanzhchen/PETuning}.}
\end{abstract}
 
\section{Introduction}
Pretrained Language Models (PLMs) such as BERT~\cite{devlin-etal-2019-bert} and RoBERTa~\cite{liu2019roberta} have orchestrated tremendous progress in NLP in the past few years, achieving state of the art on a large variety of benchmarks such as GLUE~\cite{wang-etal-2018-glue} and SuperGLUE~\cite{wang2019superglue}. Most successful applications of PLMs follow the pretraining-and-finetuning transfer learning paradigm \cite{devlin-etal-2019-bert}, where PLMs are used as backbones to be combined with additional parameters and finetuned on downstream tasks in an end-to-end manner. Whilst being simple and effective, such paradigm requires task-specific tuning of the full model that consists of hundreds of millions~\cite{devlin-etal-2019-bert, liu2019roberta}, or even billions~\cite{radford2019gpt2, brown2020language, JMLR:t5} of parameters for each task, which is time-consuming and resource-intensive. 


\begin{figure}
\centering
\includegraphics[width=\columnwidth]{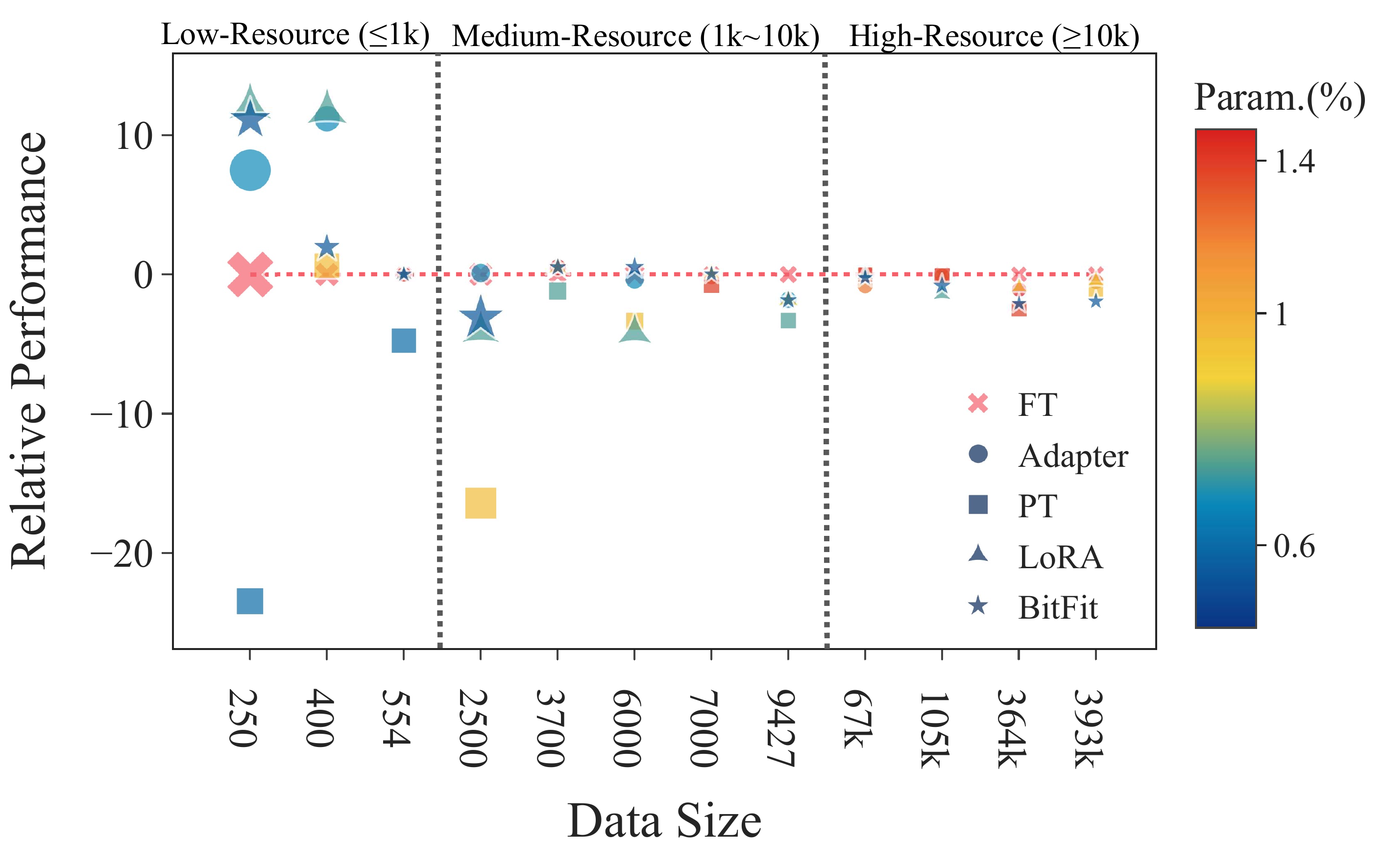}
\caption{The relative performance difference of PETuning methods, i.e., Adapter, prefix tuning (PT), LoRA, BitFit, comparing with the full finetuning (FT) over different training data size of 12 tasks from GLUE and SuperGLUE. The tasks and their split into the three resource bands are illustrated in~\Cref{sec:tasksetup}. The size of each point denotes the standard deviation and the colours of PETuning methods denote the percentage of trainable parameters over different tasks compared to full finetuning. The key takeaway message is that PETuning methods outperform finetuning only in the low-resource tasks but remain on par or behind in medium and high-resource settings. }
\label{fig:scatter}
\end{figure}

To avoid full model finetuning, there has been a surge of studies on \textbf{P}arameter-\textbf{E}fficient \textbf{Tuning} (PETuning) methods, which aim to tune the PLMs by adjusting lightweight trainable parameters while keeping most pretrained parameters frozen. 
Various ways have been used in these PETuning methods to introduce the lightweight trainable parameters. 
Adapter~\cite{houlsby2019parameterefficient, pfeiffer2020adapterhub} is one of these that injects a small portion of \textit{model-level} parameters within each transformer~\cite{NIPS2017_transformer} layer of the pretrained language model. Prompt-tuning~\cite{qin-eisner-2021-softprompt, liu2021ptuningv1,lester-etal-2021-power} is another class of methods that introduce trainable continuous embeddings into the original sequences of input token embeddings to augment the PLMs on the \textit{feature level}. Diff-pruning~\cite{guo-etal-2021-diffpruning} learns and updates additional sparse diff-vector for all pretrained parameters, and LoRA~\cite{hu2021lora} learns low-rank matrices to approximate the updated  matrices, both of which update the PLMs on the \textit{parameter level}. Moreover, BitFit~\cite{zaken2021bitfit} \textit{partially tunes} the bias terms of PLMs, without even introducing any new parameters. More details for these methods can be seen in~\Cref{section:petuning}.

Given the various exciting progresses of PETuning methods that all seem to demonstrate their competitive performance with higher training efficiency, the idea that PETuning could be a new general paradigm in place of full finetuning for transfer learning in NLP becomes never more tempting~\cite{liu2021pretrain}. We, however, argue that current evidences are insufficient to support the complete overthrow of full finetuning.
First, we point out that the current evaluation strategy, i.e., the development set is used for both early stopping and reporting results, used in a number of studies for PETuning~\cite{lester-etal-2021-power, vu2021spot, liu2021ptuning, pfeiffer2021adapterfusion} does not provide fair model comparisons.
This essentially causes data leakage that results in misleading conclusions (\Cref{section:pilot}). 
Second, statistical significance is rarely reported when comparing PETuning methods. This is an especially crucial issue as we show that the finetuning and PETuning processes are inherently unstable due to various randomness, such as weight initialization and training data order (\Cref{sec:ana_stability}).

To fairly compare these tuning strategies, this study conducts a comprehensive re-examination on the effectiveness of PETuning methods.
\textbf{Our main contributions} are:
\textbf{1)} We conduct controlled experiments (\Cref{section:pilot}) and reveal the fundamental flaw of the current evaluation scheme (i.e., its failure to assess generalisation) and how that leads to misinterpretations of the progress in the field. 
\textbf{2)} We offer a more reliable practice for model selection that is not prone to overfitting.
\textbf{3)} We revisit the performance of PETuning in comparison with finetuning across tasks with various, and have reached very different conclusions on different data scales.
\textbf{4)} We conduct the first comprehensive study to investigate the stability of off-the-shelf PETuning methods and identify the main contributing factors. 

\paragraph{Key Findings:} 
\textbf{1)} Finetuning cannot be fully replaced so far, since there is no PETuning method that can consistently outperform finetuning across all tasks and settings. We conclude that PETuning may be more suitable for low-resource tasks, but  struggle on medium-resource tasks and fall behind finetuning across the board on high-resource tasks (see Figure~\ref{fig:scatter}).
\textbf{2)} All the PETuning methods unanimously show instability across different random seeds similar to finetuning~\cite{dodge2020finetuning}, where the randomness comes from both weight initialisation and training data order. 
\textbf{3)} We found prompt-tuning lags far behind finetuning, which is a very different conclusion from previous studies. We show that prompt-tuning is highly unstable and cannot robustly and consistently re-produce its reported competitive performance (usually reported as a single run or the optimal run across multiple episodes \cite{lester-etal-2021-power, liu2021ptuning}) in our fair evaluation setup. 
\textbf{4)} Within each PETuning method, reducing the size of trainable parameters is likely to yield better stability (but not necessary to yield better or poorer performance). 
\textbf{5)} The stability of PETuning methods is substantially proportional to the scale of training data, and we further highlight the most crucial factor behind is the number of training iterations.


For the rest of the paper, we begin with the analysis on why the current evaluation protocol can be flawed (\Cref{section:pilot}), and follow with a rigorous re-examination with a fairer protocol to benchmark the performance and stability of PETuning (\Cref{section:performance}). 
\section{The Broken Protocol}

\label{section:pilot}

GLUE\footnote{\url{https://gluebenchmark.com}.} and SuperGLUE\footnote{\url{https://super.gluebenchmark.com}.} have become the \textit{de facto} benchmarks for verifying model effectiveness in Natural Language Understanding. 
For the sake of validity and fairness of the evaluation, the labels of test sets in these benchmarks are not released. Instead, web portals are provided for submitting and evaluating the prediction results.
Due to the limited number of allowed evaluation submissions to these benchmarks, a large number of works have followed a common practice that the model performance is only reported and compared based on the dev sets rather than the real test sets, where the dev set is treated as the ``test set''~\cite{lester-etal-2021-power, vu2021spot, liu2021ptuning, pfeiffer2021adapterfusion}.

While this practice is a convenient approximation of model performance as it allows quickly obtaining results from large-scaled experiments, there has been a serious data leakage problem in this setting: a single set is often used for both validating and testing the model. Therefore, the reported results under such setting might come from overly-optimistic checkpoints since early stopping is applied on the same set. 
We argue that such practice breaches the standard train/dev/test paradigm and compromises fair and rigorous comparison, leading to unreliable conclusions and misunderstandings of the examined models.

To verify the above concerns, we scrutinise the broken status-quo protocol by comparing it with a newly defined rigorous evaluation protocol. The new protocol has strictly separated sets for validation and testing. We provide comprehensive analyses to reveal the negative effects of using the dev sets for both checkpoint selection (i.e., early stopping) and testing.

\paragraph{Compared Methods.}
We have chosen four representative PETuning methods: \textbf{Adapter}, \textbf{Prompt-tuning (PT)}, \textbf{LoRA}, and \textbf{BitFit}, which correspond to model-level, feature-level, parameter-level, and partial-finetuning PETuning methods, respectively\footnote{Some works of Adapter~\cite{pfeiffer2020adapterhub} and Prompt-tuning~\cite{lester-etal-2021-power, vu2021spot, liu2021ptuning} adopt the problematic early stopping strategy (described in their experimental settings~\cite{lester-etal-2021-power, vu2021spot, pfeiffer2021adapterfusion} or code bases~\cite{liu2021ptuning}), while LoRA and BitFit adopt the standard train/dev/test paradigm.}.

For Adapter, we use the \textit{Pfeiffer} architecture~\cite{pfeiffer2020adapterhub} since it has reported better performance than others.
For Prompt-tuning, due to the poor performance of standard prompt tuning~\cite{lester-etal-2021-power} on small PLMs, e.g., base versions of Bert and RoBERTa, we adopt the settings of \textbf{prefix tuning}~\cite{li2021prefixtuning} (or P-Tuning v2~\cite{liu2021ptuning}) to add continuous prompts for each transformer layer of PLMs.
For LoRA \& BitFit, we take the architectures from their original papers~\cite{hu2021lora,zaken2021bitfit}.

\paragraph{Evaluation Setup.}
We adopt RoBERTa$_{\text{base}}$~\citep{liu2019roberta} as our base model, and experiment on the RTE dataset, which is a textual entailment dataset included in both GLUE and SuperGLUE. 
We divide the original dev set of the RTE dataset by a 50\%/50\% split\footnote{A normal way to create new dev set is to separate part of training set while using original dev set as test set, as what we do in~\Cref{section:setup}. However, to highlight the data leakage issue from the misused early stopping with a more controlled setting, we create the new dev and test sets from the same (original) dev set with similar size and distribution and fairly compare their impact for early stopping.}
(denoted by \textit{dev.1} and \textit{dev.2} respectively), and compare the performance over finetuning and the four PETuning methods.
In particular, we use the \textit{dev.2} set as the test set, and use the \textit{dev.1} set or the \textit{dev.2} set as the dev set for model selection, respectively (denoted by \textit{RTE}$_{1-2}$ or \textit{RTE}$_{2-2}$). We set the number of epochs to 50 and early stop when validation scores do not improve for 10 consecutive epochs following~\citet{mao2021unipelt}. Results will be shown for using either \textit{evaluation loss} or \textit{accuracy} as the stopping metrics.\footnote{See~\Cref{sec:hyperparameters} for the full hyperparameters settings.}

\paragraph{Results and Analyses.}
\begin{table}[!t]
    \centering
    \resizebox{\columnwidth}{!}{
    \begin {tabular}{lllll}
    \toprule
                        & \multicolumn{2}{c}{\textit{Evaluation loss}}                                              & \multicolumn{2}{c}{\textit{Accuracy}} \\                                                                                                               
        \cmidrule(r){2-3}    \cmidrule(r){4-5}             
                        & RTE$_{1-2}$                         & RTE$_{2-2}$                         &RTE$_{1-2}$                             &RTE$_{2-2}$             \\
    \midrule
            FT          & $\textbf{78.89}_{\pm 1.36}$                  & $\textbf{78.89}_{\pm 1.36}$                  & $79.28_{\pm 1.9}$                      & $\textbf{79.62}_{\pm 2.22}$       \\
            \midrule
            Adapter     & $75.1_{\pm 1.60}$                   & $\textbf{76.3}_{\pm 4.26}$                   & $76.55_{\pm 3.57}$                     & $\textbf{78.42}_{\pm 3.7}$     \\
            PT          & $57.55_{\pm 2.71}$                  & $\textbf{66.19}_{\pm 8.51}$                  & $57.84_{\pm 4.85}$                     & $\textbf{67.19}_{\pm 11.37}$    \\
            LoRA        & $75.22_{\pm 2.77}$                  & $\textbf{75.94}_{\pm 3.39}$                  & $75.11_{\pm 3.3}$                      & $\textbf{77.7}_{\pm 4.57}$      \\
            BitFit      & $70.79_{\pm 10.38}$                 & $\textbf{71.3}_{\pm 10.19}$                  & $66.76_{\pm 12.98}$                    & $\textbf{68.2}_{\pm 13.72}$      \\
    \bottomrule
    \end{tabular}}

    \caption{Mean and standard deviation results with different dev/test splits for RTE task across 20 runs. \textit{Evaluation loss} and \textit{accuracy} are the stopping metrics. Bold denotes the highest mean value for corresponding method with specific stopping metric.}
    \label{table:pilot}
\end{table}

From \Cref{table:pilot}, we can see that using a single set as both the dev and test sets (i.e. \textit{RTE}$_{2-2}$) can substantially boost the performances of PETuning models, comparing with using two separate ones (i.e., \textit{RTE}$_{1-2}$). Particularly, prefix tuning (PT) gains $\sim$10\% improvements using either evaluation loss or accuracy as the stopping metric. However, such performance boost does not mean genuine improvement in terms of better generalisation.

\begin{figure}
    \includegraphics[width=\columnwidth]{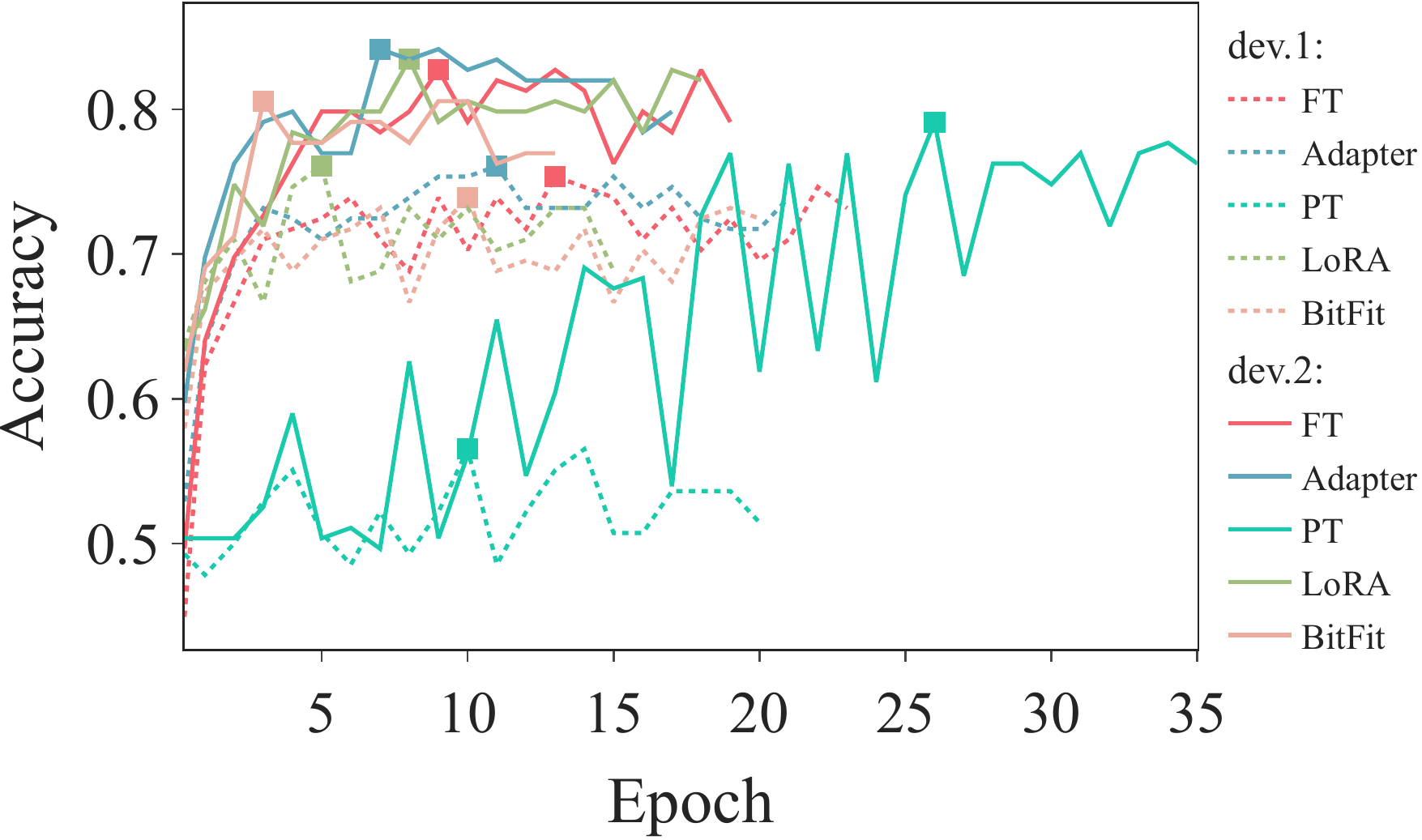}
    \caption{Comparing early stopped points selected by \textit{RTE}$_{1-2}$ and \textit{RTE}$_{2-2}$, i.e., checkpoints with the best accuracy scores from \textit{dev.1} and \textit{dev.2} over training epochs. The markers denote the epochs selected by early stopping. Comparing the two checkpoint results on dev.2 (i.e. test performance),  the \textit{RTE}$_{2-2}$ (same set for test and dev) checkpoint usually shows higher performance than the checkpoint selected in \textit{RTE}$_{1-2}$ by a large gap.}
    \label{fig:line_epoch}
\end{figure}

To demonstrate this in a more intuitive way, we plot the evaluation performance on dev sets (i.e. \textit{dev.1} and \textit{dev.2} respectively) over training steps in~\Cref{fig:line_epoch}.\footnote{The best-performing runs of \textit{RTE}$_{1-2}$ and \textit{RTE}$_{2-2}$ are used for this visualisation.} For each model, its early stopped epochs over the two sets are drastically different, suggesting that there is significant behavioural difference of the models across sets and best checkpoint selected on one set does not necessarily generalise well on the other set. 
In fact, the ability of models to mitigate such gap (e.g., from the best-performing checkpoints on \textit{dev.1} to the best-performing ones on unseen \textit{dev.2}) precisely denotes corresponding ability of generalisation, which is the most essential criteria to measure the models' effectiveness~\citep{raschka2018model}. However, the evaluation scheme \textit{RTE}$_{2-2}$, i.e., the broken protocol, reuses the test set multiple times during training stage, which is tantamount to leaking the test information to erase this gap, resulting in unreliable evaluation.

 These observations motivates us to re-examine these PETuning methods with a fairer evaluation.





\section{Experiments with Fair Evaluation}
\label{section:performance}
In this section, we use a fairer evaluation protocol that strictly separates dev and test sets. Based on this protocol, we conduct extensive experiments to investigate the effectiveness of PETuning methods (concluded in~\Cref{fig:scatter}). First, we experiment over a wide range of tasks under various levels of resource abundance to comprehensively compare the performance of PETuning with finetuning (\Cref{sec:ana_performance}). Further, we provide in-depth analyses for the instability of PETuning methods, investigating the possible causes and provide practical suggestions of using PETuning methods (\Cref{sec:ana_stability}).

\subsection{Experimental Setup}
\label{section:setup}
\begin{table*}[!t]
    \centering
        \resizebox{0.72\textwidth}{!}{
    \begin {tabular}{lccccc}
    \specialrule{0.14em}{0.5pt}{1pt} Dataset$\downarrow$, Model$\rightarrow$ & FT                                    &Adapter                                          &PT                                      &LoRA                                        &BitFit                 \\
    
    \specialrule{0.14em}{0.5pt}{0.5pt}
    \multicolumn{6}{c}{\textit{Low-Resource}}\\
    \specialrule{0.06em}{0.5pt}{1pt}
            CB          & $70.00_{\pm 13.32}$                             & $77.49_{\pm 13.20}$                             & $46.55_{\pm 5.74}^{\downarrow}$                   & $\textbf{82.05}_{\pm 9.62}^{\uparrow}$                         & $81.12_{\pm 8.94}^{\uparrow}$     \\
            COPA        & $54.70_{\pm 3.36}$                              & $65.90_{\pm 5.42}^{\uparrow}$                              & $55.35_{\pm 5.07}$           & $\textbf{66.4}_{\pm 9.05}^{\uparrow}$                          & $56.65_{\pm 3.72}$     \\
            WSC         & $\textbf{63.46}_{\pm 0.0}$                               & $\textbf{63.46}_{\pm 0.0}$                               & $58.7_{\pm 4.69}^{\downarrow}$                    & $\textbf{63.46}_{\pm 0.0}$                          & $\textbf{63.46}_{\pm 0.0}$      \\
    \textbf{Avg. (Low)}       & $62.72_{\pm 4.58}$                              & $68.95_{\pm 4.83}^{\uparrow}$                              & $53.53_{\pm 3.22}^{\downarrow}$           & $\textbf{70.64}_{\pm 4.32}^{\uparrow}$                         & $67.08_{\pm 3.57}^{\uparrow}$      \\
    \specialrule{0.14em}{0.5pt}{0.5pt}
    \multicolumn{6}{c}{\textit{Medium-Resource}}\\
    \specialrule{0.06em}{0.5pt}{1pt}    
            RTE         & $73.77_{\pm 3.17}$                              & $\textbf{73.88}_{\pm 1.88}$                              & $57.36_{\pm 8.01}^{\downarrow}$                                & $69.69_{\pm 7.89}^{\downarrow}$                         & $70.67_{\pm 10.77}$    \\
            MRPC        & $90.54_{\pm 1.05}$                              & $\textbf{91.06}_{\pm 0.63}$                              & $89.35_{\pm 1.31}^{\downarrow}$                                & $91.03_{\pm 0.95}$                         & $\textbf{91.06}_{\pm 0.71}$     \\
            WiC         & $65.47_{\pm 2.04}$                              & $65.12_{\pm 1.88}$                              & $62.12_{\pm 1.32}^{\downarrow}$                                & $61.29_{\pm 6.7}^{\downarrow}$                          & $\textbf{66.0}_{\pm 1.41}$      \\
            STS-B       & $90.42_{\pm 0.26}$                              & $90.23_{\pm 0.1}^{\downarrow}$                               & $89.64_{\pm 0.39}$                                & $\textbf{90.47}_{\pm 0.11}$                         & $90.44_{\pm 0.15}$                  \\
            BoolQ        & $\textbf{78.75}_{\pm 0.72}$                              & $76.93_{\pm 0.92}$                              & $75.44_{\pm 0.47}$                                & $76.92_{\pm 1.33}$                         & $76.9_{\pm 0.84}$                  \\
    \textbf{Avg. (Medium)}       & $\textbf{79.79}_{\pm 0.99}$                              & $79.44_{\pm 0.74}$                              & $74.78_{\pm 1.62}^{\downarrow}$                                & $77.88_{\pm 2.02}^{\downarrow}$                        & $79.01_{\pm 2.22}$\\
    \specialrule{0.14em}{0.5pt}{0.5pt}
    \multicolumn{6}{c}{\textit{High-Resource}}\\
    \specialrule{0.06em}{0.5pt}{1pt}    
            SST-2       & $\textbf{94.15}_{\pm 0.0}$                               & $93.34_{\pm 0.31}^{\downarrow}$                              & $\textbf{94.15}_{\pm 0.0}$                                 & $\textbf{94.15}_{\pm 0.0}$                & $93.92_{\pm 0.07}$                                                     \\
            QNLI        & $\textbf{92.40}_{\pm 0.12}$                              & $92.31_{\pm 0.09}$                              & $92.31_{\pm 0.27}$                                & $91.00_{\pm 0.69}$               & $91.60_{\pm 1.01}$                                                     \\
            QQP         & $\textbf{91.38}_{\pm 0.06}$                              & $90.28_{\pm 0.0}$                               & $88.90_{\pm 0.32}^{\downarrow}$                                & $90.45_{\pm 0.17}^{\downarrow}$               & $89.28_{\pm 0.0}^{\downarrow}$                                                     \\
            MNLI        & $\textbf{87.42}_{\pm 0.20}$                              & $86.88_{\pm 0.17}^{\downarrow}$                              & $86.30_{\pm 0.08}^{\downarrow}$                                & $86.96_{\pm 0.24}$               & $85.50_{\pm 0.32}^{\downarrow}$                                                   \\
    \textbf{Avg. (High)}       & $\textbf{91.34}_{\pm 0.09}$                              & $90.70_{\pm 0.12}^{\downarrow}$                 & $90.42_{\pm 0.14}^{\downarrow}$                                & $90.64_{\pm 0.21}^{\downarrow}$               & $90.08_{\pm 0.29}^{\downarrow}$\\
    \specialrule{0.14em}{1pt}{1pt}
    \textbf{Avg. (All)}  & $79.37$                                         & $\textbf{80.57}$                                         & $74.68$                                           & $80.32$                          & $79.72$\\
    \specialrule{0.14em}{1pt}{1pt}             
    \end{tabular}}
    \caption{Mean and standard deviation results for each of the 12 tasks across finetuning (FT) and four PETuning methods.
    We report the F1 score for CB and MRPC, Pearson correlation for STS-B, and accuracy for other tasks (matched accuracy for MNLI). Higher is better for all metrics. One-tailed t-test is used for the comparison between PETuning and finetuning.  One PETuning method outperforms ($\uparrow$) or falls behind ($\downarrow$) finetuning when accepting the corresponding alternative hypothesis, where p-value < 0.05 (meaning the difference is significant).
    }
    \label{table:overall_results}
\end{table*}



\paragraph{Data Setup.}
We conduct experiments on 12 datasets from GLUE and SuperGLUE, which are divided into three levels according to their sizes: (1) \textit{low-resource} (< 1k data points), including CB~\citep{deMarneffe_Simons_Tonhauser_2019_CB}, COPA~\citep{roemmele2011choice_COPA}, and WSC~\citep{10.5555/3031843.3031909_WSC}; (2) \textit{medium-resource} (1k \textasciitilde 10k data points), including RTE~\citep{wang-etal-2018-glue}, MRPC~\citep{dolan2005automatically_MRPC}, WiC~\citep{pilehvar-camacho-collados-2019-WIC}, STS-B~\citep{cer-etal-2017-semeval_STS-B}, and BoolQ~\citep{clark-etal-2019-boolq}; (3) \textit{high-resource} (> 10k data points), including SST-2~\citep{wang-etal-2018-glue}, MNLI~\citep{williams-etal-2018-broad_MNLI}, QNLI~\citep{wang-etal-2018-glue}, and QQP\footnote{\url{https://quoradata.quora.com/First-Quora-Dataset-Release-Question-Pairs}}.

Since using a single set for both early stopping and testing could result in unreliable results (\Cref{section:pilot}), we use separate dev and test sets for all our experiments. Specifically, the original training set of each dataset is split into new train set and dev set by a 90\%/10\% proportion, and the original dev set is used as the test set.\footnote{Ideally, the standard train-dev-test splits of GLUE and SuperGLUE should be used. However, due to the amount of experiments and evaluations need to be done in our ultra-large-scale investigation, we create our own splits instead of submitting models to the learderboards.} 

\paragraph{Evaluation Setup.}
For the aforementioned four PETuning methods, i.e., Adapter, prefix tuning, LoRA, and BitFit,
we again experiment with the RoBERTa$_{\text{base}}$ model~\citep{liu2019roberta} on our 12 datasets. All experimental results are reported across 20 runs for low- and medium-resource tasks, and 10 runs for high-resource tasks with different random seeds, respectively.
We train for 50 epochs and early stop the training when evaluation loss does not decrease for 10 consecutive epochs.\footnote{See~\Cref{sec:hyperparameters} for the full hyperparameters settings.}




\subsection{Analysis of Performance}
\label{sec:ana_performance}
From the average performance for all tasks in \Cref{table:overall_results}, we can observe that most of the PETuning methods (i.e., Adapter, LoRA, and BitFit) indeed have some performance gains when compared with finetuning.
It is known that PETuning methods have far better tuning efficiency, with significantly less tuning parameters ($<2\%$ of full model parameters), comparing with full finetuning~\citep{mao2021unipelt}.
However, it remains questionable whether
PETuning methods are more advantageous as the overall comparison may neglect important divergences in the wide range of tasks with different scales of training data.
To provide a finer-grained view for the comparison between finetuning and PETuning, we group the results of the 12 tasks in~\Cref{table:overall_results} into low-, medium-, and high-resource tasks.
Whilst most PETuning methods outperform finetuning on low-resource settings, the best PETuning is merely comparable to finetuning in medium-resource tasks and lags behind finetuning in high-resource tasks. We summarise the trend in~\Cref{table:trend}, and provide more detailed analyses in the following.

\begin{table}[!t]
    \centering
    \resizebox{0.75\columnwidth}{!}{
    \begin{tabular}{l ccc}
    \toprule
                    & Low      & Medium   & High \\
    \midrule
    Adapter         & $\nearrow$                  &  $\longrightarrow$                           &  $\searrow$             \\
    \specialrule{0em}{1pt}{1pt}
    PT              & $\searrow$                  &  $\searrow$                            &  $\searrow$             \\
    LoRA            & $\nearrow$                  &  $\searrow$                  & $\searrow$              \\
    BitFit          & $\nearrow$                  &  $\longrightarrow$                 & $\searrow$              \\
    \bottomrule
    \end{tabular}}
    \caption{Performance comparison between PETuning and finetuning on low-, medium-, and high-resource settings, respectively. Arrows indicate whether corresponding PETuning method significantly outperforms finetuning ($\nearrow$),  falls behind ($\searrow$), or their results across multiple runs without significant differences ($\rightarrow$).
    }
    \label{table:trend}
\end{table}

\begin{figure}[!t]
    \includegraphics[width=\columnwidth]{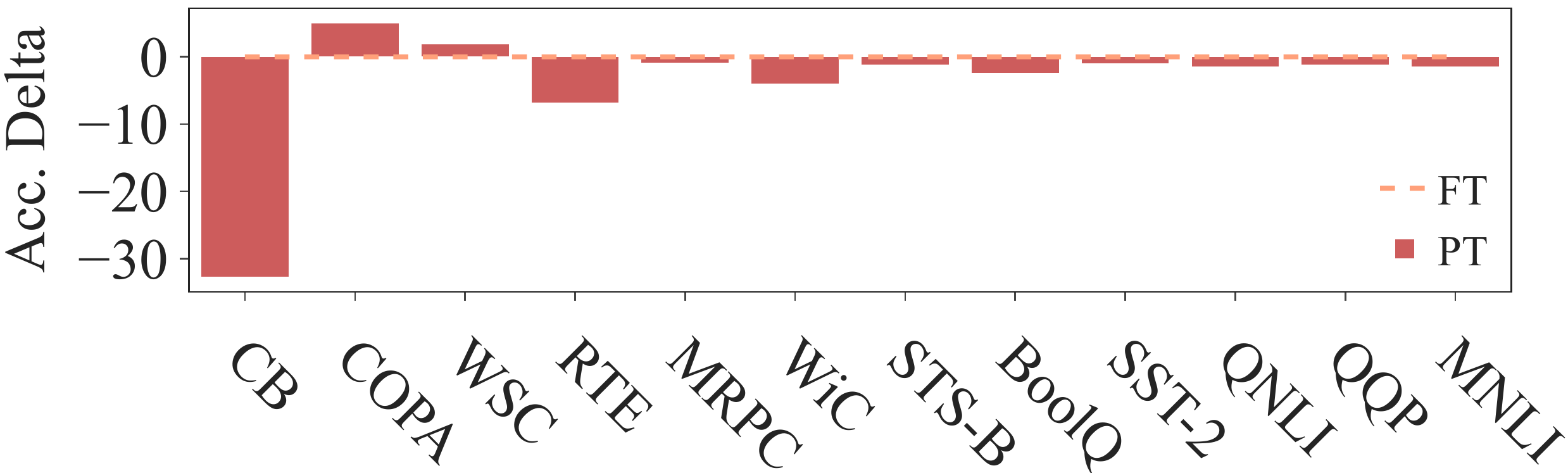}
    \caption{Relative performance differences of prefix tuning (PT) over full finetuning (FT) on the upper bounds of multi-run results. PT achieves close upper bounds compared with FT on most of the 12 tasks.}
    \label{fig:prompt_upper}
\end{figure}

\begin{figure*}[!t]
    \centering
    \includegraphics[width=0.9\textwidth]{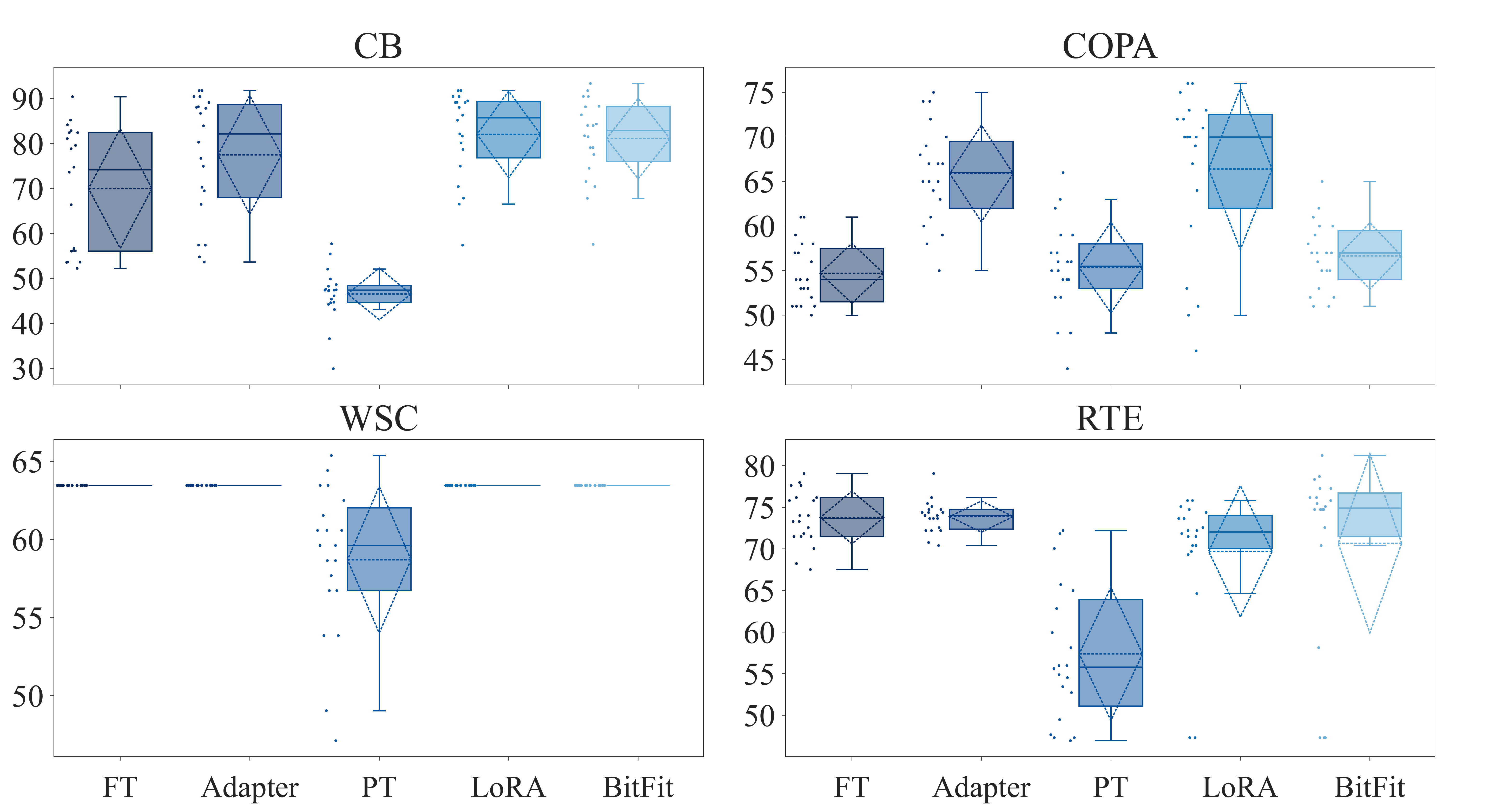}
    \caption{The experimental results over 20 different random seeds across CB, COPA, WSC, and RTE datasets, where finetuning and PETuning methods show large instability. The dashed rhombuses denote the mean (horizontal dashed line) and standard deviation (vertical distance).}
    \label{fig:overall_box}
\end{figure*}

\paragraph{\textit{Adapter \& LoRA \& BitFit} only perform better on low-resource tasks.}
From~\Cref{table:overall_results}, we observe that Adapter, LoRA, and BitFit obtain outstanding performance on the low-resource tasks and significantly outperform finetuning by large margins\footnote{Similar observation for Adapter was previously reported in \citet{he-etal-2021-effectiveness}. We extend it to more PETuning methods.} (especially LoRA obtains \textasciitilde8\% performance gains on average).
However, the trend changes when training data size gets larger. For the medium-resource tasks, only Adapter and BitFit can maintain a comparable performance with finetuning. LoRA and prefix tuning lags behind substantially. For the high-resource setting, finetuning performs consistently better than all PETuning methods.\footnote{These findings are also observed on the same task with different number of training instances. See ~\Cref{sec:diff_datasize} for more details.} In particular, among the PETuning methods, Adapter obtains the highest scores on high-resource tasks. These results suggest that low-resource is the only setting where PETuning methods could outperform full finetuning.

\paragraph{\textit{Prefix tuning} consistently underperforms finetuning.}
According to \Cref{table:overall_results} and \Cref{table:trend}, finetuning beats prefix tuning by large margins on most tasks across multiple runs, contradicting to what has been reported in \citet{liu2021ptuning}. One possible reason is that prefix tuning is highly unstable to train and thus may have exploited the broken protocol more than other PETuning methods (see \Cref{fig:line_epoch}). Besides using a flawed evaluation protocol, previous works on prefix tuning only report their result of a single run \citep{liu2021ptuning,lester-etal-2021-power, vu2021spot}, which might lead to biased conclusion.
In \Cref{fig:prompt_upper} we further plot the upper bounds of these runs, and we indeed observe that the optimal run from prefix tuning achieves competitive performance compared with finetuning on many tasks. However, the results in~\Cref{table:overall_results} verify that this competitiveness would plummet across different runs by varying the random seeds.
Such instability of prefix tuning leads to its poor average performance in our experiments. We further discuss this in~\Cref{sec:ana_stability}.


\paragraph{Finetuning cannot be fully replaced.}
To summarise, PETuning has exceptional performance in resource-poor scenarios and usually outperform the more expensive full-model finetuning. However, when dataset size increases, finetuning regains dominance in medium- and high-resource setups. This indicates that finetuning cannot be fully replaced so far.
We also delved deeper into understanding why finetuning lags behind PETuning on low-resource settings. Our investigation points to the different fitting capabilities of finetuning and PETuning. Specifically, finetuning is more prone to overfitting on low-resource tasks.\footnote{See~\Cref{sec:low_finetuning} for more details and analyses.}



\begin{table}[!t]
    \centering
        \resizebox{\columnwidth}{!}{
    \begin {tabular}{lccc}
    \toprule
    & WI                                      &DO                                        &Global                 \\
    
    \midrule
            FT                               & $\textbf{55.40}_{\pm 4.55}$        &  $55.35_{\pm \textbf{3.32}}$           &$54.70_{\pm 3.36}$     \\
            \midrule
            Adapter                          & $\textbf{67.15}_{\pm \textbf{5.40}}$        & $66.35_{\pm 7.36}$            & $65.90_{\pm 5.42}$    \\
            PT                               & $55.00_{\pm 5.13}$        & $54.75_{\pm \textbf{4.97}}$            & $\textbf{55.35}_{\pm 5.07}$    \\
            LoRA                             & $63.60_{\pm \textbf{7.93}}$        & $64.60_{\pm 8.56}$            & $\textbf{66.40}_{\pm 9.05}$    \\
            BitFit                           & $\textbf{58.40}_{\pm \textbf{2.29}}$        & $56.00_{\pm 4.00}$            & $56.65_{\pm 3.72}$    \\
   \bottomrule
    \end{tabular}}
    \caption{Performance over 20 runs on COPA task, controlled by global random seeds, weight initialization (WI) random seeds, and data order (DO) random seeds, respectively. (Visualised in~\Cref{fig:wi_do} in the Appendix.)}

    \label{table:wi_do}
\end{table}

    


\begin{figure*}[!t]
    \includegraphics[width=0.95\textwidth]{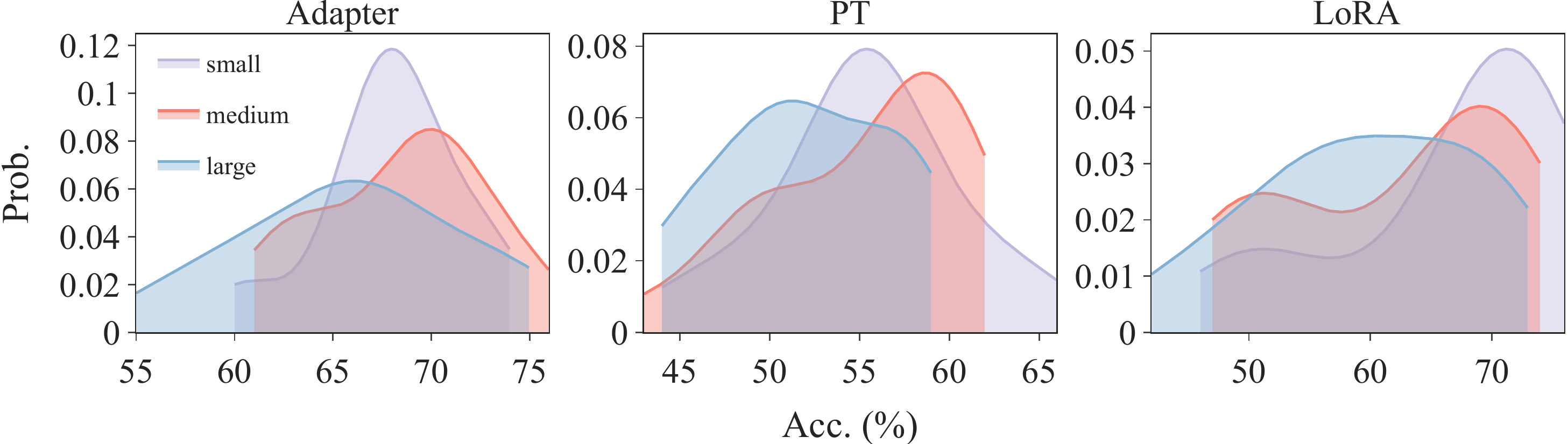}
    \caption{Performance probability density curves of Adapter, prefix tuning (PT), and LoRA over small, medium, and large parameter scales on COPA task across 20 runs. (See the numerical results and analyses in~\Cref{sec:parameter_ana}.)}
    \label{fig:parameter}
\end{figure*}

\subsection{Analysis of Stability}
\label{sec:ana_stability}

By revisiting the results in ~\Cref{table:overall_results}, we can observe that both finetuning and all PETuning methods exhibit large standard deviations on several tasks, i.e., CB, COPA, WSC, and RTE.
To further understand this phenomenon, in~\Cref{fig:overall_box}, we visualise the performance distribution of 20 runs of finetuning and PETuning methods on these tasks. Surprisingly, large fluctuations are seen on all four tasks across all methods, where the margins between the lower and upper bounds could reach over 30\%. While \citet{dodge2020finetuning, conf/iclr/MosbachAK21} have previously identified such variation exists for finetuning, our experiments further validate that such instability also occurs in all PETuning methods and could be even more prominent in certain tasks.

This level of instability severely hampers the application of PETuning and there is a pressing need to understand the underlying cause. However, to the best of our knowledge, no previous studies have systematically discussed the instability issue in PETuning methods. In this section, we provide the first comprehensive investigation on this matter. While instability is measured as the performance differences between random seeds, we further disentangle two randomness sources (weight initialisation and training data order) to better describe model instability. We then investigate two factors that might affect model instability: (1) trainable parameter size; and (2) training data size and training iterations. Through controlled experiments, we find that model instability is reflected by both changing data order and changing weight initialisation. Reducing model size and increasing training iteration seems to have positive impact on model stability. We discuss all these points in detail in the followings.

\paragraph{Weight initialisation and data order work together.}
Instability is measured from performance changes due to randomness introduced by random seeds.
Two key things impacted by random seeds are (a) the initialisation of trainable weights (including extra parameters of PETuning methods and the classification head), and (b) the order of training data fed to the model. To disentangle these two factors, following the setting in~\citet{dodge2020finetuning}, we use two separate random seeds to control weight initialisation and training data order respectively, comparing with using one global random seed to control these two factors simultaneously.

\Cref{table:wi_do} demonstrates that each of the two factors could individually lead to large standard deviations, which means the instability of PETuning methods are sensitive to either training data order, or weight initialisation, or both. 
This observation indicates that the sources of instability for PETuning can be multifaceted -- isolating and enhancing stability via controlling individual factor can be challenging.\footnote{Prior works mainly focused on obtaining better prior (e.g., prompt/weight initialisation) to improve model performance/stability but did not touch upon the multifaceted nature of instability \citep{pfeiffer2021adapterfusion,lester-etal-2021-power, vu2021spot}.}

\paragraph{Models with fewer trainable parameters are more stable.}

To investigate the impact of model size on model stability, we define three sizes, \textit{small}, \textit{medium}, and \textit{large}, for each PETuning method. The three sizes correspond to the reduction factor of \{64, 16, 2\} for Adapter\footnote{The smaller the reduction factor, the more parameters the model has.}, the prompt length of \{32, 64, 128\} for prefix tuning, and the rank of \{8, 16, 32\} for LoRA.
We conduct a set of controlled experiments on the COPA task where PETuning methods exhibit high instability.
We perform 20 runs for each setting and use kernel density estimation (KDE)~\citep{kde} to estimate the probability density curves of the multi-run results.

As shown in ~\Cref{fig:parameter}, for all PETuning methods, we consistently observe that the probability density curves would be progressively flatter (having lower peak) as the number of parameters increase from small to large. This suggests that more trainable parameters for PETuning leads to a wider range of performance distribution, resulting in higher instability.
That said, when it comes to model performance, the best-performing model usually is not the smallest one. 
We conjecture that models with fewer trainable parameters converge quickly to the rough global minima but could be underfitting the real data manifold. 

\begin{figure}[!t]
    \centering
    \includegraphics[width=\columnwidth]{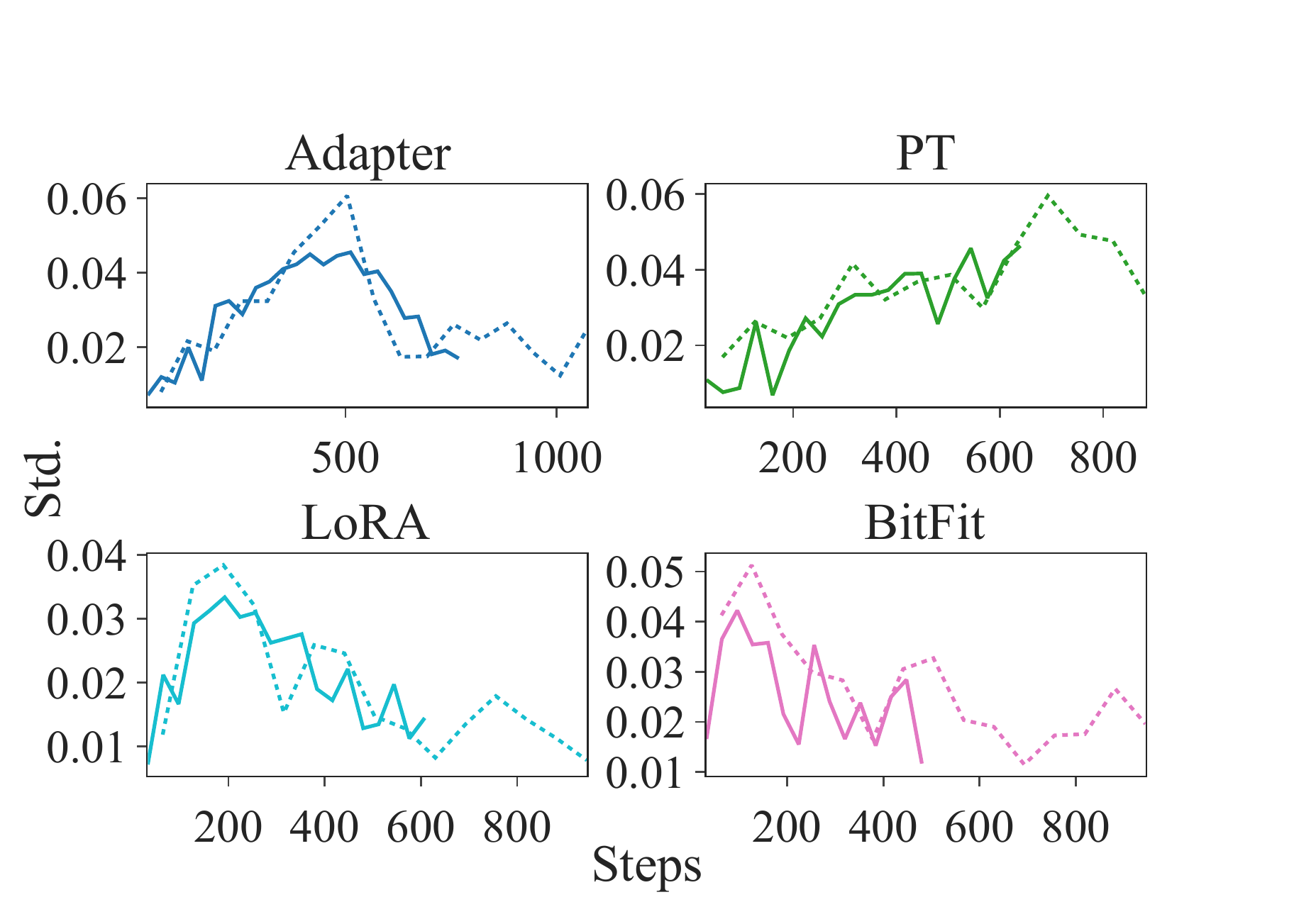}
    \caption{Standard deviations of data size in \{1k (solid line), 2k (dashed line)\} over training steps on WiC task across 20 runs. (See that on BoolQ task in~\Cref{fig:dataline_boolq} in the Appendix.)}
    \label{fig:dataline}
\end{figure}

\paragraph{Data size does not affect instability directly, but training steps do.}
\Cref{fig:scatter} and ~\Cref{table:overall_results} suggest that PETuning methods almost always have larger standard deviations on lower-resource tasks.\footnote{This is further confirmed in~\Cref{sec:diff_datasize}.}
To investigate if training data size directly affects the stability of PETuning, inspired by ~\citet{conf/iclr/MosbachAK21}, we compare models that are trained with randomly sampled 1k and 2k training instances from WiC training set and validated with another separately sampled 1k dev set.

In ~\Cref{fig:dataline}, we observe that the solid and dashed lines of each PETuning method are substantially intertwined, which means the standard deviations (instability) of PETuning methods trained by 1k or 2k samples would not have significant differences with the same number of steps. The true underlying variable that leads to the discrepancy of instability across different training data sizes is essentially the number of training iterations/steps. As shown in~\Cref{fig:dataline}, the standard deviations of PETuning methods have an initial ascent stage where models are fitting to the training data and thus having fluctuating performance. After the ascent stage,
the standard deviations substantially decrease as the number of training steps get larger. 
With number of epochs being fixed, the total number of iterations on small datasets is small and the standard deviation has yet to decrease, causing the higher instability in lower-resource tasks.
In particular, due to the weaker fitting capabilities~\citep{deltatuning}, prefix tuning (PT) has a longer ascent stage, which might need more training iterations to obtain more stable performance.
That said, prolonging the training on small datasets does not necessarily enhance model performance, and the best checkpoint may still be only appearing when the standard deviation is high.


\section{Related Work}

\paragraph{Instability of finetuning PLMs.} 
While our study is, to the best of our knowledge, the first to systematically investigate PETuning instability, prior studies have looked into the instability of finetuning PLMs.
\citet{dodge2020finetuning} illustrated the inherent instability of finetuning by controlling random seeds and provided a new early stopping strategy to improve instability.
\citet{LeeCK20mixout} proposed a new regularisation method by mixing two models based on dropout to prevent catastrophic forgetting and to improve instability.
More recently, \citet{conf/iclr/MosbachAK21} revisited the hypotheses of finetuning instability proposed by previous studies and found that optimisation difficulties can lead to vanishing gradients, which further causes finetuning instability. \citet{zhang2021revisiting} also revealed that optimisation significantly affects the instabilities in few-sample fine-tuning.

\paragraph{Analysis of PETuning.}
As PETuning methods have become a prominent research direction, a great number of studies aim to analyse the characteristics of these methods.
\citet{he-etal-2021-effectiveness} investigated the effectiveness of Adapter across different scales and \citet{han-etal-2021-robust} provided a robust strategy for training Adapter.
Recently, \citet{he2021unified} and \citet{mao2021unipelt} proposed a unified view to connect various PETuning methods. However, there has not been reliable validation and comparison for off-the-shelf PETuning methods in terms of stability and effectiveness, and this is where our paper bridges the gap.
\section{Conclusion}
This work conducted a rigorous re-examination on the current Parameter-Efficient Tuning (PETuning) methods. We demonstrated that performing early stopping and evaluation on the same dataset (a common practice used in many past studies) could lead to unreliable conclusions. This issue is more pronounced when accompanied by the instability nature of PETuning, leading to inflated results and overly optimistic estimates of PETuning approaches. 
We re-evaluated these PETuning methods on the performance and stability aspects on
a rigorous evaluation protocol that strictly separates validation and test sets. By conducting a set of fine-grained comparisons between PETuning and finetuning, we found that PETuning methods are not consistently competitive with finetuning. Namely, prefix tuning performs poorly across tasks and most PETuning methods perform worse than finetuning on higher-resource settings. By systematically investigating the instability of PETuning methods, we found that models' instability is sensitive to both weight initialisation and training data order. We identify two major factors behind such instability: 1) models with fewer parameters are more stable within each PETuning method class; 2) more training iterations can usually reduce instability. Our overall re-examination conclude that finetuning still cannot be fully replaced by PETuning so far, and there are many key challenges for PETuning in terms of both performance and instability, which need to be addressed in future work.

\section*{Limitations}
This work provides a comprehensive study and analysis for the existing popular PETuning methods, i.e., Adapter, Prompt-Tuning (prefix tuning), LoRA, and BitFit, focusing on their performance and stability.
Empirically, we use standard deviations to measure the stability of these PETuning methods across multiple runs. Standard deviation is more reliable when having more number of runs. A larger number of runs would contribute to more precise estimation of such stability. 
We chose 20 runs for low- and medium-resource tasks and 10 runs for high-resource tasks. However, larger numbers of runs can consolidate our conclusions.

Besides, we used the available train and dev sets from GLUE and SuperGLUE to simulate a standard train/dev/test split. The conclusion would be more comparable to existing works if having access to the real testing data.

Last but not least, we covered four representative PETuning methods. However, PETuning is a fast-moving field and our conclusions do not necessarily generalise to all existing and upcoming models.
\section*{Acknowledgements}
This research was partially supported by the National Natural Science Foundation of China (Grant No. 61906219),
and the Mohamed bin Zayed University of Artificial Intelligence, United Arab Emirates.

\bibliography{anthology,custom}
\bibliographystyle{acl_natbib}

\appendix

\appendix
\clearpage
\section*{Appendix}

\section{PETuning Methods}
\label{section:petuning}
PETuning methods are unique in keeping (most) pretrained parameters of PLMs frozen and finetuning only light-weight additional parameters or a fraction of the PLM's parameters for downstream tasks.

\begin{figure}[!t]
\centering
\includegraphics[width=0.8\columnwidth]{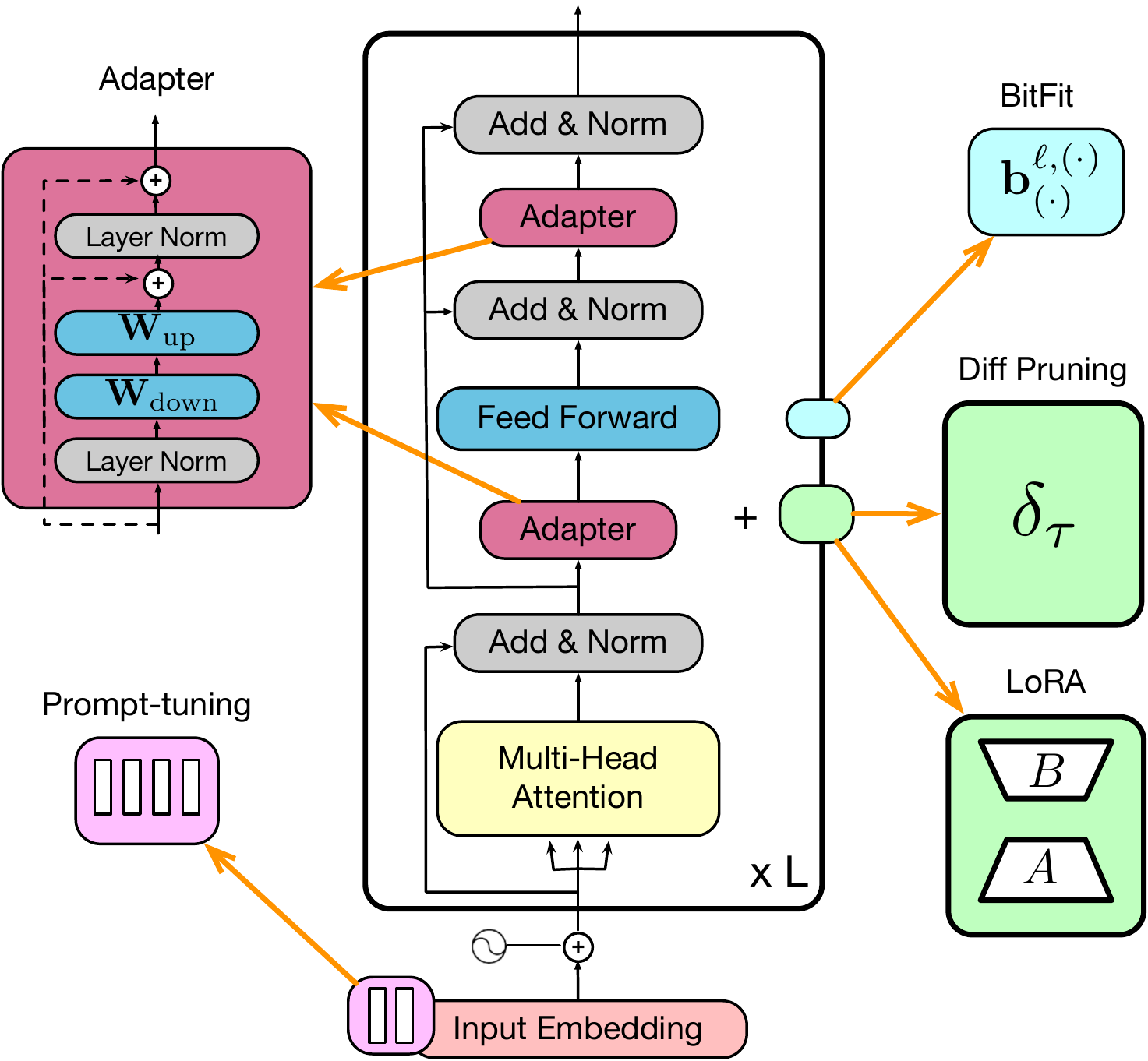}
\caption{Different PETuning methods by adjusting trainable parameter on model level (Adapter), feature level (Prompt-tuning), parameter level (Diff Pruning and LoRA), and partial-tuning level (BitFit).}
\label{fig:PETuning}
\end{figure}

To achieve efficient tuning of PLMs, existing PETuning methods are generally designed by two different manners: (1) training additional parameters on different levels of PLMs, including model-level (\Cref{sec:model_level}), feature-level (\Cref{sec:feature_level}), and the parameter-level (\Cref{sec:param_level}), or (2) tuning partial parameters of the base model (\Cref{sec:partial_finetune}). \Cref{fig:PETuning} shows the difference of these PETuning methods.
\subsection{Model-Level }\label{sec:model_level}

\paragraph{Adapter-Tuning.}
\textit{Adapters}~\cite{houlsby2019parameterefficient,pfeiffer2020adapterhub,pfeiffer2021adapterfusion, meng-etal-2021-mixture} are a type of PETuning approaches that insert small newly initialised parameter modules on the model-level (i.e., each transformer layer) of PLMs. In particular, these adapter modules are normally moulded by a two-layer feed-forward neural network with a bottleneck: (1) a down-projection with $\mathbf{W}_{\text{down}} \in \mathbb{R}^{d \times r}$ to project the input $\textbf{h}_i$ to a lower-dimensional space specified by bottleneck dimension $r$;
(2) an up-projection with $\mathbf{W}_{\text{up}} \in \mathbb{R}^{r\times d}$ to project back to the input size. Mathematically, the adapter can be defined as:
\begin{equation}
    \mathbf{h}_a = \mathbf{W}_{\text{up}}^{\top} f \left(\mathbf{W}_{\text{down}}^{\top}\mathbf{h}_i\right),
\end{equation}
where $\mathbf{h}_a$ is the output and $f(\cdot)$ is the activation function. During the finetuning, the model only updates the parameters of the adapter modules while keeps the underlying pretrained model fixed.
\subsection{Feature-Level}\label{sec:feature_level}

\paragraph{Prompt-Tuning.}
Prompt-Tuning~\cite{lester-etal-2021-power} is another type of PETuning approaches that introduce additional tunable parameters on the feature-level.
Specifically, prompt-tuning introduces additional tunable prefix (or suffix) vectors, namely prompts~\citep{zhong-etal-2021-factual, schick-schutze-2021-just}, to extend the input text features (or the input of each transformer layer~\cite{li2021prefixtuning,liu2021ptuning}), and tunes only the prompts. Besides its simplicity and lightness, prompt-tuning could achieve  on par performance, particularly in billions-size PLMs, and even better performance, comparing with the full finetuning~\cite{liu2021ptuning}.

\subsection{Parameter-Level}\label{sec:param_level}

\paragraph{Diff-Pruning.}
Diff-pruning~\cite{guo-etal-2021-diffpruning} works on all parameters of PLMs, which aims to learn additional trainable sparse parameters for the entire PLMs. Specifically, for the pretrained parameters $\Theta$, diff-pruning reparameterizes the task-specific model parameters $\Theta_{\tau}$ as:
\begin{equation}
    \Theta_{\tau} = \Theta + \delta_{\tau},
\end{equation}
where $\delta_{\tau}$ denotes the trainable diff vector, which is regularised to be sparse.
\paragraph{LoRA.}
LoRA~\cite{hu2021lora} focuses on the updating procedure of the language model parameters. For a pretrained weight matrix $\mathbf{W} \in \mathbb{R}^{d\times k}$, LoRA uses trainable low-rank matrices to approximate the updates ($\Delta \mathbf{W}$) by:
\begin{equation}
    \mathbf{W} + \Delta \mathbf{W} = \mathbf{W} + \mathbf{B}\mathbf{A},
\end{equation}
where $\mathbf{B} \in \mathbb{R}^{d\times r}, \mathbf{A} \in \mathbb{R}^{r\times k}$, and the rank $r \ll \min(d, k)$.

\subsection{Partial Finetuning}\label{sec:partial_finetune}

\paragraph{BitFit.}
\textit{Partial finetuning} aims to tune a fraction of PLMs parameters without introducing any additional ones. For example, \citet{lee2019elsa} only tunes the top layers, however, which usually performs much worse than full finetuning.
With the principle of efficiency and effectiveness, BitFit~\cite{zaken2021bitfit} turns to tune the bias terms of PLMs to obtain competitive performance.

\section{General Experimental Setup}
In this section, we illustrate the general task and hyperparameter settings. Apart from that, in~\Cref{section:pilot} and ~\Cref{section:performance}, we will additionally illustrate their specific data and evaluation setups, respectively.
\label{section:gsetup}
\subsection{Task Setup}
\label{sec:tasksetup}
In order to extensively compare the performance and stability of PETuning methods with the full-model finetuning, we select a full set of 12 tasks across low-, medium- and high-resource scales of GLUE and SuperGLUE,
including natural language inference (CB, RTE, MNLI, QNLI), question answering (COPA, BoolQ), paraphrasing (MRPC, QQP), 
sentiment analysis (SST-2), sentence similarity (STS-B), word sense disambiguation (WiC), and coreference resolution (WSC) tasks. 
According to the dataset sizes, we divide these tasks into three levels:
\begin{itemize}
    \item \textbf{Low-Resource:} the tasks with training data size smaller than 1k, including CB, COPA, and WSC.
    \item \textbf{Medium-Resource:} the tasks with training data size between 1k and 10k, including RTE, MRPC, WiC, STS-B, and BoolQ.
    \item \textbf{High-Resource:} the tasks with training data size larger than 10k, including SST-2, QNLI, QQP, and MNLI.
\end{itemize}


\subsection{Hyperparameter Setup}
\label{sec:hyperparameters}
We adopt Roberta$_{\text{base}}$ as the base model released by Huggingface\footnote{\url{https://github.com/huggingface/transformers}}.
The grid search is used to select the learning rate from \{1e-6, 1e-5, 5e-5, 1e-4, 5e-4, 1e-3, 5e-3, 1e-2\} and batch size from \{16, 32\}.
We search the reduction factor from \{2, 16, 64\} following~\cite{pfeiffer2021adapterfusion} for Adapter, the prompt length from \{8, 16, 32, 64\} for prefix tuning, and the scaling factor $\alpha$ and rank from \{8, 16\} for LoRA following its origin paper.
There are many studies focusing on achieving better initialization by post pretraining for PETuning methods such as Adapter~\cite{pfeiffer2021adapterfusion} and prompt~\cite{vu2021spot, gu2021ppt}, however,
to be a fair comparison, the extra parameters of all PETuning methods are initialized randomly.

We set the number of epochs to 50 and adopt the early stopping strategy with the patience of 10 worse-performing epochs on our new development set following~\cite{mao2021unipelt}. In particular, for ~\Cref{section:pilot}, to fully investigate the effects of early stopping on the task RTE, we use both \textit{evaluation loss} and \textit{accuracy} as the stopping metrics; for~\Cref{section:performance}, due to the variety of evaluation metrics for the tasks, we use the \textit{evaluation loss} as the common stopping metric.


\section{Additional Experiments and Analyses}

\subsection{The Same Task with Different Training Data Sizes}
\label{sec:diff_datasize}
\begin{table*}[!t]
    \centering
    \resizebox{\textwidth}{!}{
    \begin {tabular}{lcccccccc}
        \toprule Model$\downarrow$, Dataset$\rightarrow$
                        & WiC                              &STS-B                                &BoolQ                                    &SST-2                               &QNLI                                   &QQP                       &MNLI                          &Avg.\\
        \specialrule{0.1em}{0.5pt}{0.5pt} 
                \multicolumn{9}{c}{\textit{500}} \\
        \specialrule{0.1em}{0.5pt}{1pt}                                               
            FT          & $56.12_{\pm 2.13}$              & $83.81_{\pm 1.34}$                  & $62.17_{\pm 0.0}$                        & $87.89_{\pm 1.42}$                 & $77.54_{\pm 6.5}$                     & $74.21_{\pm 3.11}$       & $58.61_{\pm 6.21}$           & $71.47_{\pm 1.44}$  \\
            Adapter     & $58.36_{\pm 3.98}^{\uparrow}$   & $85.74_{\pm 0.64}^{\uparrow}$       & $61.56_{\pm 1.51}$                       & $89.12_{\pm 0.84}^{\uparrow}$      & $79.85_{\pm 1.55}^{\uparrow}$         & $75.91_{\pm 1.01}^{\uparrow}$       & $60.92_{\pm 2.86}^{\uparrow}$           & $73.07_{\pm 0.84}^{\uparrow}$  \\
            PT          & $56.25_{\pm 1.07}$              & $73.98_{\pm 3.79}^{\downarrow}$     & $57.55_{\pm 4.43}^{\downarrow}$          & $82.01_{\pm 2.75}^{\downarrow}$    & $70.77_{\pm 5.16}^{\downarrow}$       & $66.37_{\pm 1.42}^{\downarrow}$       & $36.35_{\pm 1.51}^{\downarrow}$           & $63.32_{\pm 1.28}^{\downarrow}$  \\
            LoRA        & $58.64_{\pm 1.16}^{\uparrow}$   & $85.06_{\pm 0.83}^{\uparrow}$       & $60.48_{\pm 4.74}$                       & $89.1_{\pm 0.78}^{\uparrow}$       & $80.86_{\pm 0.7}^{\uparrow}$                     & $72.43_{\pm 2.9}$        & $63.1_{\pm 2.57}^{\uparrow}$            & $72.81_{\pm 0.84}^{\uparrow}$  \\
            BitFit      & $57.92_{\pm 2.4}^{\uparrow}$    & $84.39_{\pm 1.98}^{\uparrow}$       & $62.16_{\pm 0.04}$                       & $88.38_{\pm 8.19}$                 & $78.38_{\pm 2.74}$                    & $73.13_{\pm 1.63}$       & $61.47_{\pm 2.08}^{\uparrow}$           & $72.26_{\pm 1.23}^{\uparrow}$  \\
        \specialrule{0.1em}{0.5pt}{0.5pt}
                \multicolumn{9}{c}{\textit{5k}} \\
        \specialrule{0.1em}{0.5pt}{1pt}    
            FT          & $66.98_{\pm 1.26}$               & $90.76_{\pm 0.03}$                  & $73.81_{\pm 1.11}$                      & $92.66_{\pm 0.92}$          & $87.12_{\pm 0.37}$          & $83.72_{\pm 0.17}$       & $78.18_{\pm 0.82}$         & $81.89_{\pm 0.44}$ \\
            Adapter     & $65.15_{\pm 2.85}$               & $90.24_{\pm 0.05}$                  & $73.51_{\pm 1.63}$                      & $92.66_{\pm 0.25}$          & $86.69_{\pm 0.49}$          & $82.97_{\pm 0.32}$       & $78.0_{\pm 1.17}$          & $81.32_{\pm 0.71}$\\
            PT          & $66.25_{\pm 1.29}$               & $89.14_{\pm 0.66}$                  & $62.32_{\pm 0.14}^{\downarrow}$                      & $91.7_{\pm 0.92}$           & $85.83_{\pm 1.32}^{\downarrow}$          & $80.11_{\pm 1.52}^{\downarrow}$       & $77.78_{\pm 0.42}$         & $79.02_{\pm 0.83}^{\downarrow}$\\
            LoRA        & $62.46_{\pm 1.38}^{\downarrow}$  & $90.57_{\pm 0.15}$                  & $71.9_{\pm 0.41}^{\downarrow}$                       & $92.35_{\pm 0.7}$           & $87.49_{\pm 0.39}$          & $83.03_{\pm 0.36}$       & $77.67_{\pm 0.26}$         & $80.78_{\pm 0.45}^{\downarrow}$\\
            BitFit      & $67.61_{\pm 0.49}$               & $90.37_{\pm 0.19}$                  & $75.08_{\pm 0.56}^{\uparrow}$                      & $92.35_{\pm 0.56}$          & $86.47_{\pm 0.28}$          & $82.9_{\pm 0.25}$        & $78.87_{\pm 0.1}$          & $81.95_{\pm 0.15}$\\
        \specialrule{0.1em}{0.5pt}{0.5pt}
     \multicolumn{9}{c}{\textit{50k}} \\
        \specialrule{0.1em}{0.5pt}{1pt}    
            FT          & -               & -                  & -                      & $93.46_{\pm 0.18}$          & $90.07_{\pm 0.22}$          & $88.36_{\pm 0.18}$       & $84.71_{\pm 0.41}$                                         & $89.15_{\pm 0.27}$\\
            Adapter     & -               & -                  & -                      & $93.02_{\pm 0.25}$          & $89.03_{\pm 0.17}^{\downarrow}$          & $86.67_{\pm 0.26}^{\downarrow}$       & $84.03_{\pm 0.56}$               & $88.19_{\pm 0.18}^{\downarrow}$\\
            PT          & -               & -                  & -                      & $93.32_{\pm 0.47}$          & $88.23_{\pm 0.59}^{\downarrow}$          & $85.21_{\pm 0.79}^{\downarrow}$       & $82.96_{\pm 0.20}^{\downarrow}$  & $87.43_{\pm 0.41}^{\downarrow}$\\
            LoRA        & -               & -                  & -                      & $93.35_{\pm 0.27}$          & $89.49_{\pm 0.13}$          & $87.20_{\pm 0.33}^{\downarrow}$       & $83.26_{\pm 0.11}^{\downarrow}$               & $88.33_{\pm 0.20}^{\downarrow}$\\
            BitFit      & -               & -                  & -                      & $92.99_{\pm 0.38}$          & $89.00_{\pm 0.09}^{\downarrow}$          & $87.51_{\pm 0.14}^{\downarrow}$       & $83.25_{\pm 0.08}^{\downarrow}$  & $88.19_{\pm 0.12}^{\downarrow}$\\
        \bottomrule             
    \end{tabular}}
    \caption{Mean and standard deviation results for the 7 tasks by 500, 5k, and 5k samples of training data sets across 20 runs.}
    \label{table:resources}
\end{table*}
To make the above conclusions in~\Cref{sec:ana_performance} more convincing, we conduct fine-grained experiments following~\cite{he-etal-2021-effectiveness}.
Specifically, we separately sample 500, 5k and 50k training instances from the original training data as representatives of low-, medium- and high-resource settings, in addition to draw another 1k samples as development set for each task.
We report experimental results for WiC, STS-B, BoolQ, SST-2, QNLI, QQP, and MNLI, which have more than 6k training samples, and following the settings illustrated in \Cref{section:setup}.

Confirming our conclusions, in~\Cref{table:resources}, we obtain fully consistent findings with~\Cref{sec:ana_performance} and ~\Cref{table:trend}, that prefix tuning consistently falls behind finetuning on various-resources tasks; Adapter\&LoRA\&BitFit significantly outperforms finetuning on low-resource tasks; Adapter\&BitFit keep competitive with finetuning and LoRA lags behind; and all PETuning methods falls behind on high-resource tasks.

\begin{figure}[!t]
\centering
    \includegraphics[width=\columnwidth]{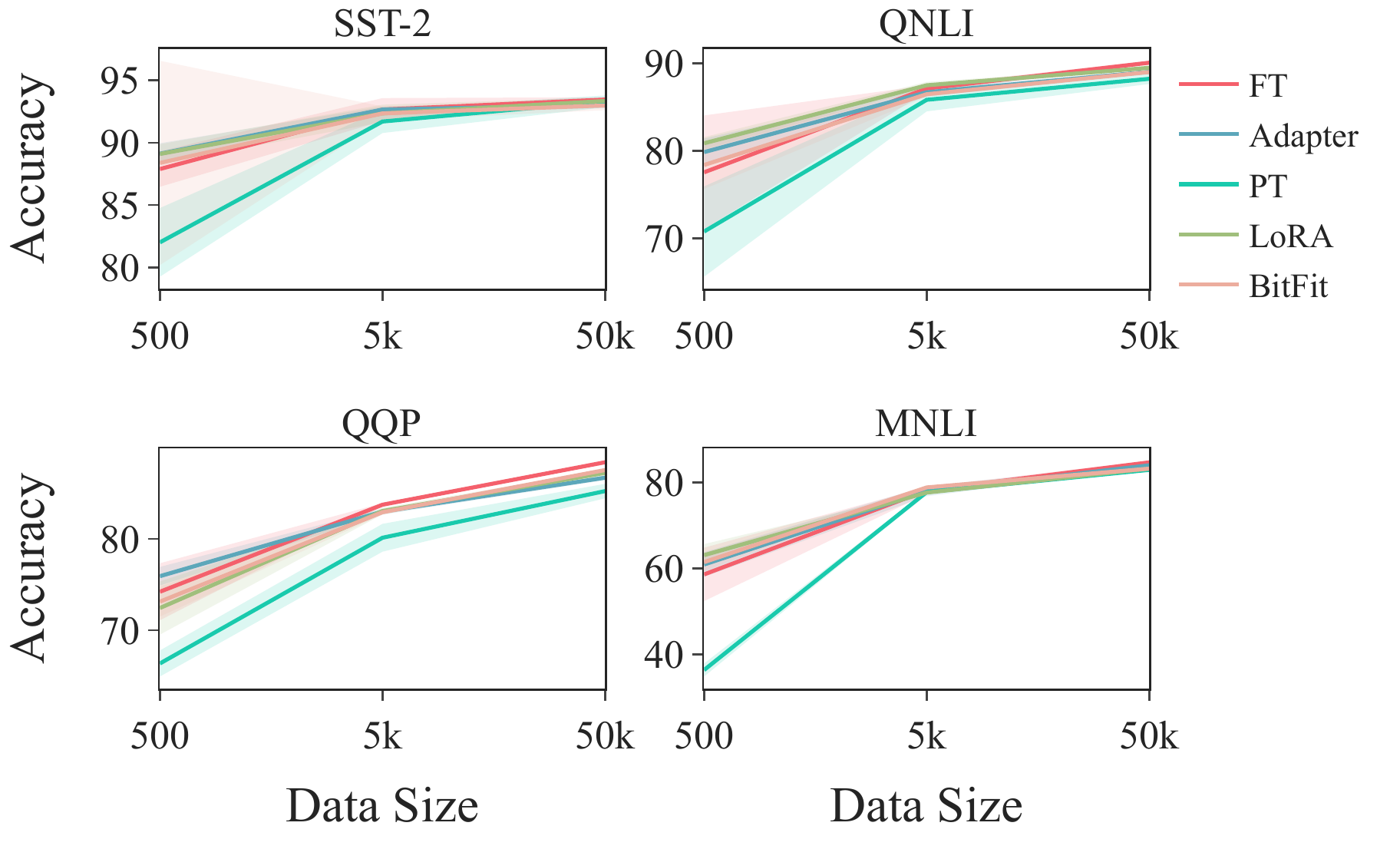}
    \caption{Performance over data scale in {500, 5k, 50k} on SST-2, QNLI, QQP, and MNLI. The shaded regions are the standard deviations.}
    \label{fig:datasize}
\end{figure}

In addition, we plot the mean and std. values with different data scales on the same task in~\Cref{fig:datasize}, to confirm the std. is substantially proportional to training data size.

\subsection{Finetuning is More Prone to Overfit}
\label{sec:low_finetuning}
\begin{figure}[!t]
    \includegraphics[width=0.95\columnwidth]{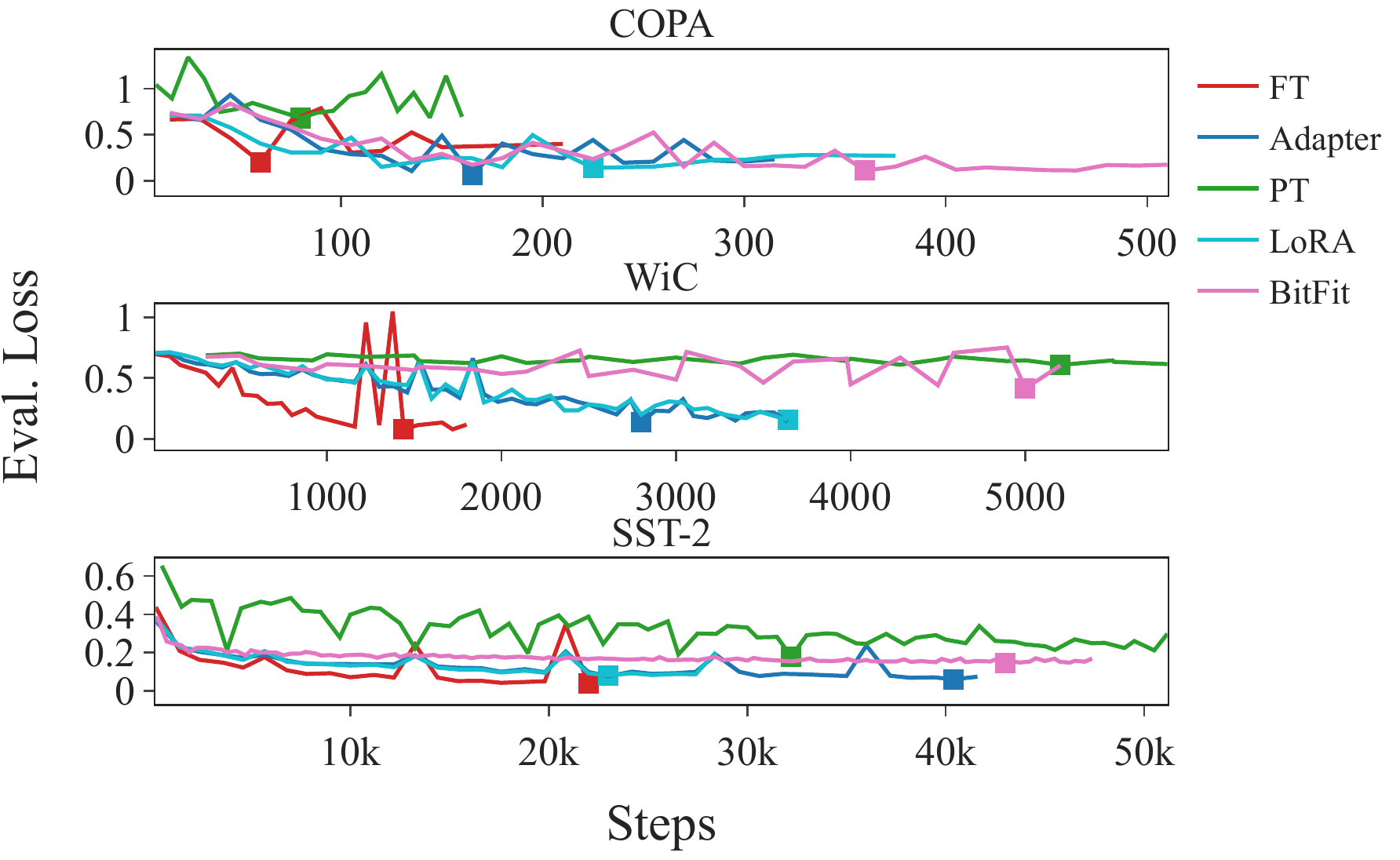}
    \caption{Evaluation loss over training steps on COPA, WiC, and SST-2.}
    \label{fig:eval_loss_line}
\end{figure}
To investigate the reasons behind the performance discrepancy of finetuning and PETuning under different training resources, we plot the evaluation loss over training steps for COPA, WiC, and SST-2, as the representatives of low-, medium-, and high-resource tasks, respectively in~\Cref{fig:eval_loss_line}. We observe that finetuning always converges faster than PETuning, especially on low-resource task COPA, where the training steps are less than 100. One possible explanation for the aforementioned discrepancy is that finetuning could converge faster than PETuning methods, which might cause the overfitting issue on low-resource settings, subsequently leading to the poorer performance.
\begin{figure*}[!t]
    \includegraphics[width=\textwidth]{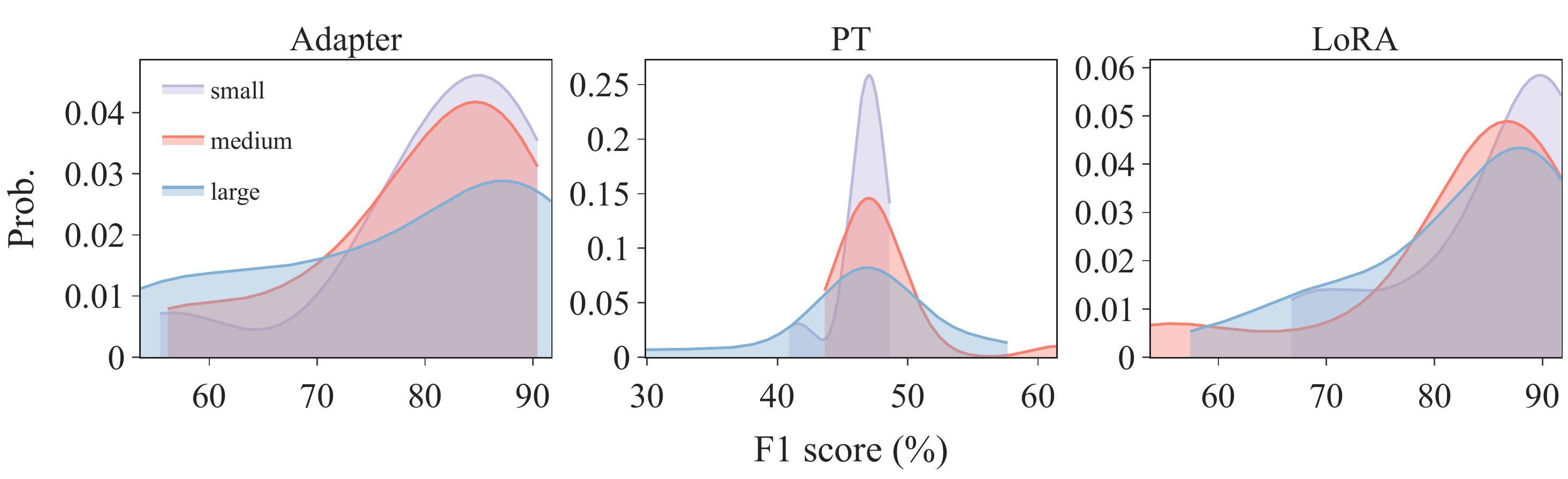}
    \caption{Performance probability density curves of Adapter, prefix tuning (PT), and LoRA over small, medium, and large parameter scales on CB task across 20 runs.}
    \label{fig:parameter_cb}
\end{figure*}
\begin{figure*}[!t]
    \includegraphics[width=0.95\textwidth]{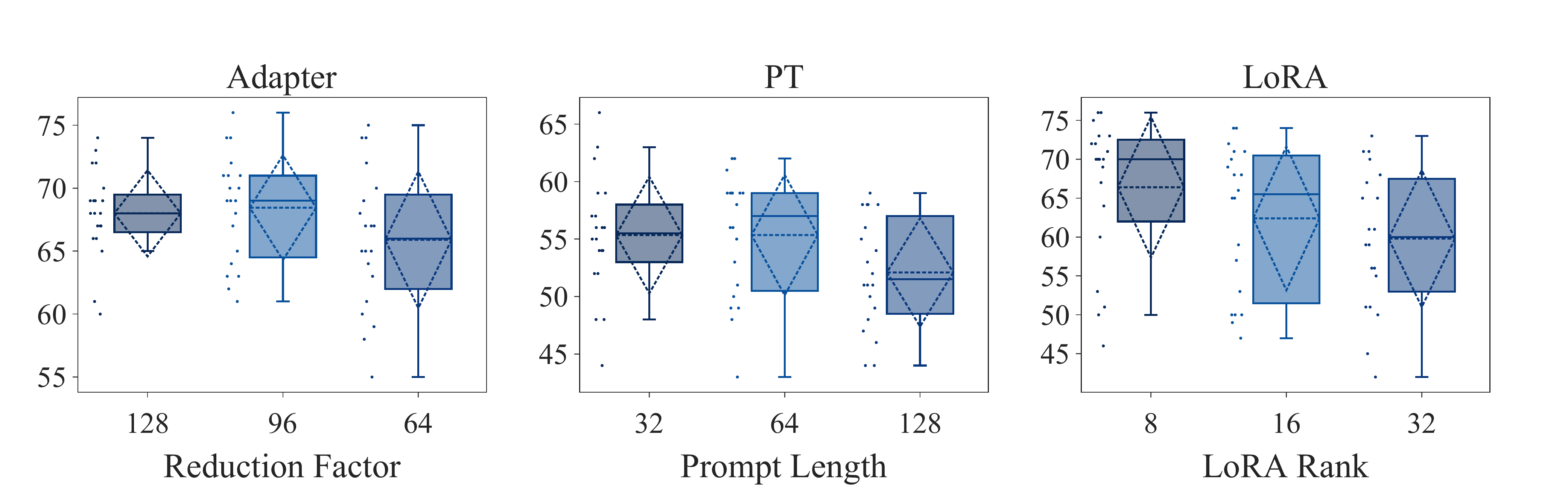}
    \caption{Performance over Adapter, prefix tuning (PT), and LoRA over small, medium, and large parameter scales on COPA task.}
    \label{fig:box_parameter}
\end{figure*}
\begin{table}[!t]
    \centering
        \resizebox{\columnwidth}{!}{
    \begin {tabular}{lccc}
    \specialrule{0.12em}{0.5pt}{1pt}                                                                           &Small                                      &Medium                                        &Large                 \\
    
    \specialrule{0.1em}{0.5pt}{1pt}
            Adapter                               & $68.0_{\pm \textbf{3.42}}$        &  $\textbf{68.45}_{\pm 4.17}$          &$65.9_{\pm 5.42}$     \\
            PT                          & $\textbf{55.35}_{\pm \textbf{4.71}}$        & $\textbf{55.35}_{\pm 5.07}$            & $52.1_{\pm 5.24}$    \\
            LoRA                               & $\textbf{66.4}_{\pm 9.05}$       & $62.4_{\pm \textbf{8.99}}$            & $59.8_{\pm 9.24}$    \\
    \specialrule{0.1em}{1pt}{1pt}             
    \end{tabular}}
    \caption{Performance over 20 runs on COPA task, controlled by global random seeds, weight initialization (WI) random seeds, and data order (DO) random seeds, respectively.}

    \label{table:different_parameters}
\end{table}

\begin{figure}[!t]
    \includegraphics[width=\columnwidth]{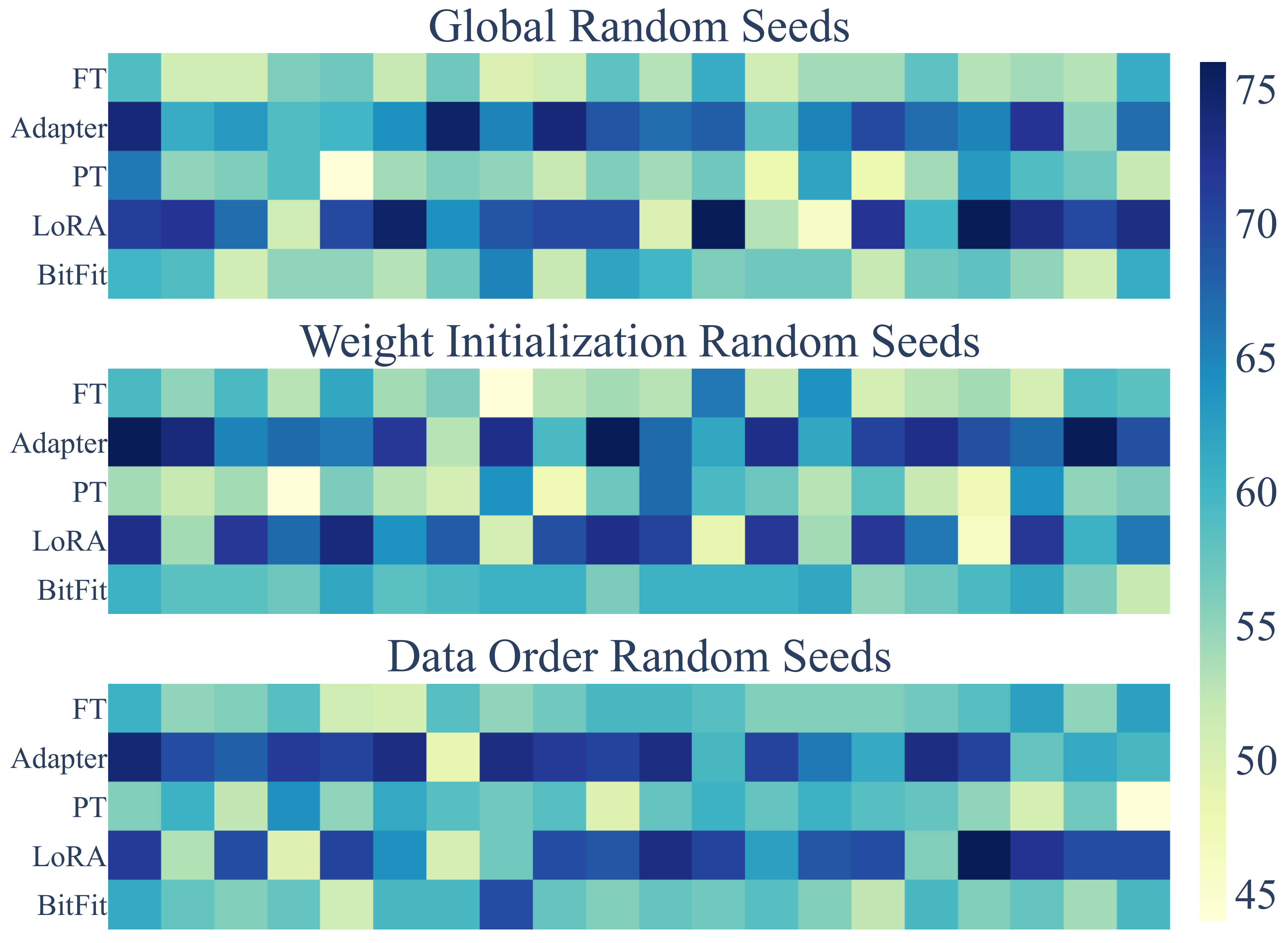}
    \caption{Performance over 20 runs on RTE, controlled by global random seeds, weight initialization (WI) random seeds, and data order (DO) random seeds, respectively.}
    \label{fig:wi_do}
\end{figure}

\begin{figure}[!t]
    \centering
    \includegraphics[width=\columnwidth]{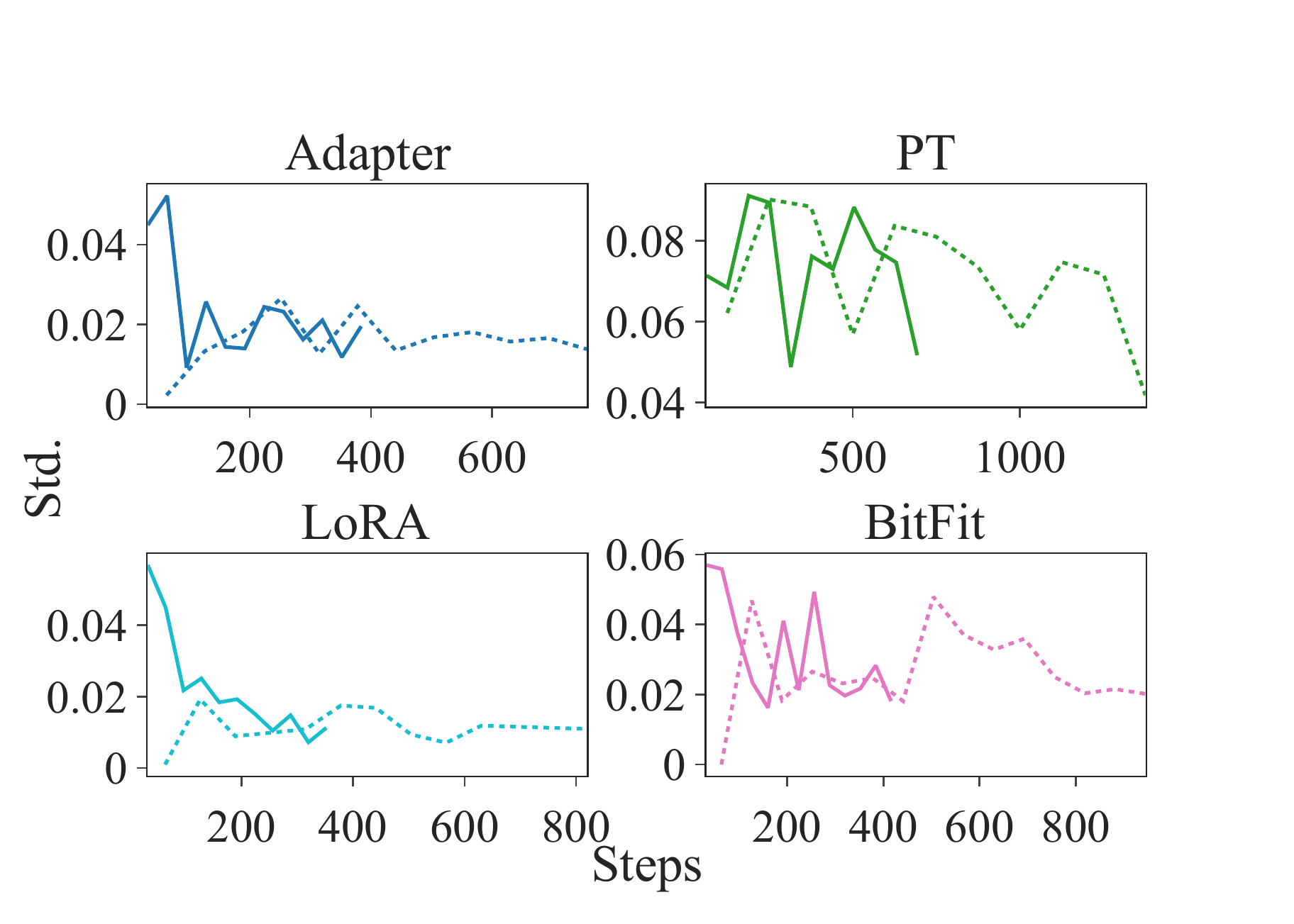}
    \caption{Standard deviations of data size in \{1k (solid line), 2k (dashed line)\} over training steps on BoolQ task across 20 runs.}
    \label{fig:dataline_boolq}
\end{figure}
\subsection{High Stability on Fewer Trainable Parameters.}
\label{sec:parameter_ana}
The probability density curves (\Cref{fig:parameter} and~\Cref{fig:parameter_cb}) have statistically confirmed PETuning methods tend to exhibit higher stability with fewer trainable parameters. In~\Cref{table:different_parameters}, we also directly list the numerical results of Adapter, PT, and LoRA over small, medium, and large parameter scales across 20 runs. While the results substantially support our conclusion that Adapter and PT achieve lowest standard deviations on the small parameter scale, except LoRA obtains slightly lower std. on the medium one. To gain more understanding about the multi-run results, we visualise them in~\Cref{fig:box_parameter}. Confirming our conclusion, we can observe that PETuning methods indeed show a trend towards clustering points and smaller boxes on small parameter scale, which means probably higher stability. However, there are also likely to generalise outliers on small parameter scale as shown in~\Cref{fig:box_parameter}, especially under our limited 20 runs, which could lead to the increasing variance. This special case might result in the inconsistent phenomenon of LoRA with other PETuning methods, nonetheless, the results and phenomena in~\Cref{table:different_parameters} and~\Cref{fig:box_parameter} generally further support the conclusion that PETuning are likely to have high stability on fewer trainable parameters.

\end{document}